\documentclass[11pt]{article}
\usepackage{acl2001,times}
\title{{\bf Learning Computational Grammars}}

\author{
John Nerbonne\thanks{~University of Groningen, 
     \{nerbonne,konstant\}@let. \hspace*{0.25cm} rug.nl, 
     osborne@cogsci.ed.ac.uk}~,
Anja Belz\thanks{~SRI Cambridge, anja.belz@cam.sri.com,
     Rob.Koe- \hspace*{0.25cm} ling@netdecisions.co.uk}~, 
Nicola Cancedda\thanks{~~XRCE Grenoble, nicola.cancedda@xrce.xerox.com}~,
{\bf Herv\'e D\'ejean}\thanks{~University of T\"ubingen,
     Herve.Dejean@xrce.xerox.  \hspace*{0.25cm} com,
     thollard@sfs.nphil.uni-tuebingen.de}~,\\
{\bf James Hammerton}\thanks{~~University College Dublin, 
     james.hammerton@ucd.ie}~~,
{\bf Rob Koeling}$^{\dag}$,
{\bf Stasinos Konstantopoulos}$^{*}$,\\
{\bf Miles Osborne}$^{*}$,
{\bf Franck Thollard}$^{\S}$ \and
{\bf Erik Tjong Kim Sang\thanks{~~University of Antwerp, erikt@uia.ua.ac.be}}
}

\date{\today}

\begin{document}

\maketitle

\vspace*{-0.0cm}
\begin{abstract}
\noindent
This paper reports on the {\sc Learning Computational Grammars} (LCG)
project, a postdoc network devoted to studying the application of
machine learning techniques to grammars suitable for computational
use.  We were interested in a more systematic survey to understand the
relevance of many factors to the success of learning, esp. the
availability of annotated data, the kind of dependencies in the data,
and the availability of knowledge bases (grammars).  We focused on
syntax, esp. noun phrase (NP) syntax.
\end{abstract}

\section{Introduction}

This paper reports on the still preliminary, but already satisfying
results of the {\sc Learning Computational Grammars} (LCG) project, a
postdoc network devoted to studying the application of machine
learning techniques to grammars suitable for computational use.  The
member institutes are listed with the authors and also included ISSCO
at the University of Geneva.  We were impressed by early experiments
applying learning to natural language, but dissatisfied with the
concentration on a few techniques from the very rich area of machine
learning.  We were interested in a more systematic survey to
understand the relevance of many factors to the success of learning,
esp. the availability of annotated data, the kind of dependencies in
the data, and the availability of knowledge bases (grammars).  We
focused on syntax, esp. noun phrase (NP) syntax from the beginning.
The industrial partner, Xerox, focused on more immediate applications
\cite{Cancedda-Samuelsson-00b}.

The network was focused not only by its scientific goal, the
application and evaluation of machine-learning techniques as used to
learn natural language syntax, and by the subarea of syntax chosen, NP
syntax, but also by the use of shared training and test material, in
this case material drawn from the Penn Treebank.  Finally, we were
curious about the possibility of combining different techniques,
including those from statistical and symbolic machine learning.  
The network members played an important role in the organisation of
three open workshops in which several external groups participated,
sharing data and test materials.

\section{Method}

This section starts with a description of the three tasks that we have
worked on in the framework of this project.
After this we will describe the machine learning
algorithms applied to this data and conclude with some notes about
combining different system results.

\subsection{Task descriptions}

In the framework of this project, we have worked on the following three
tasks:

\begin{enumerate}
\itemsep -0.2cm
\item base phrase (chunk) identification
\item base noun phrase recognition
\item finding arbitrary noun phrases
\end{enumerate}

\noindent
Text chunks are non-overlapping phrases which contain syntactically
related words.
For example, the sentence:

\begin{quote}
$[_{\rm NP}$ He $]$ 
$[_{\rm VP}$ reckons $]$ 
$[_{\rm NP}$ the current account deficit $]$ 
$[_{\rm VP}$ will narrow $]$
\\
$[_{\rm PP}$ to $]$ 
$[_{\rm NP}$ only $\pounds$ 1.8 billion $]$ 
\\
$[_{\rm PP}$ in $]$ 
$[_{\rm NP}$ September $]$ 
.
\end{quote}

\noindent
contains eight chunks, four NP chunks, two VP chunks and two PP
chunks.
The latter only contain prepositions rather than prepositions plus the
noun phrase material because that has already been included in NP
chunks.
The process of finding these phrases is called {\sc chunking}.
The project provided a data set for this task at the CoNLL-2000
workshop \cite{tks2000c}\footnote{
Detailed information about chunking, the CoNLL-2000 shared task, is
also available at http://lcg-www.uia.ac.be/conll2000/chunking/}.
It consists of sections 15-18 of the Wall Street Journal part of
the Penn Treebank II \cite{marcus93} as training data (211727 tokens)
and section 20 as test data (47377 tokens).

A specialised version of the chunking task is {\sc NP chunking} or
baseNP identification in which the goal is to identify the base noun
phrases.
The first work on this topic was done back in the eighties 
\cite{church88}.
The data set that has become standard for evaluation machine learning
approaches is the one first used by Ramshaw and Marcus
\shortcite{ramshaw95}.
It consists of the same training and test data segments of the Penn
Treebank as the chunking task (respectively sections 15-18 and section
20).
However, since the data sets have been generated with different
software, the NP boundaries in the NP chunking data sets are slightly
different from the NP boundaries in the general chunking data.

Noun phrases are not restricted to the base levels of parse trees.
For example, in the sentence {\it In early trading in Hong Kong Monday
, gold was quoted at \$ 366.50 an ounce .}, the noun phrase 
$[_{\rm NP}$ \$ 366.50 an ounce $]$ contains two embedded noun phrases
$[_{\rm NP}$ \$ 366.50 $]$ and $[_{\rm NP}$ an ounce $]$.
In the {\sc NP bracketing} task, the goal is to find all noun phrases
in a sentence.
Data sets for this task were defined for CoNLL-99\footnote{
Information about NP bracketing can be found at
http://lcg-www.uia.ac.be/conll99/npb/
}.
The data consist of the same segments of the Penn Treebank as the previous
two tasks (sections 15-18) as training material and section 20 as test
material.
This material was extracted directly from the Treebank and therefore
the NP boundaries at base levels are different from those in the
previous two tasks. 

In the evaluation of all three tasks, the accuracy of the learners
is measured with three rates.  We compare the constituents postulated
by the learners with those marked as correct by experts (gold standard).
First, the percentage of detected constituents that are correct
(precision).
Second, the percentage of correct constituents that are detected (recall).
And third, a combination of precision and recall, the F$_{\beta=1}$
rate which is equal to (2*precision*recall)/(precision+recall).

\subsection{Machine Learning Techniques}

This section introduces the ten learning methods that have been
applied by the project members to the three tasks:
LSCGs, ALLiS, LSOMMBL, Maximum Entropy, Aleph, MDL-based DCG learners,
Finite State Transducers, {\sc ib1ig}, {\sc IGTree} and C5.0.

{\bf Local Structural Context Grammars} ({\sc lscg}s)
\cite{belz2001} are situated between conventional probabilistic context-free
production rule grammars and {\sc dop}-Grammars (e.g.,\ Bod and Scha
\shortcite{bod97}).  {\sc lscg}s outperform the former because they do
not share their inherent independence assumptions, and are more
computationally efficient than the latter, because they incorporate
only subsets of the context included in {\sc dop}-Grammars.  Local
Structural Context ({\sc lsc}) is (partial) information about the
immediate neighbourhood of a phrase in a parse.  By conditioning
bracketing probabilities on {\sc lsc}, more fine-grained probability
distributions can be achieved, and parsing performance increased.

Given corpora of parsed text such as the {\sc wsj}, {\sc lscg}s are
used in automatic grammar construction as follows.  An {\sc lscg} is
derived from the corpus by extracting production rules from
bracketings and annotating the rules with the type(s) of {\sc lsc} to
be incorporated in the {\sc lscg} (e.g.\ parent category information,
depth of embedding, etc.).  Rule probabilities are derived from rule
frequencies (currently by Maximum Likelihood Estimation).  In a
separate optimisation step, the resulting {\sc lscg}s are optimised in
terms of size and parsing performance for a given parsing task by an
automatic method (currently a version of beam search) that searches
the space of partitions of a grammar's set of nonterminals.

The LSCG research efforts differ from other approaches reported in
this paper in two respects.  Firstly, no lexical information is used
at any point, as the aim is to investigate the upper limit of parsing
performance without lexicalisation.  Secondly, grammars are optimised
for parsing performance {\em and} size, the aim being to improve
performance but not at the price of arbitrary increases in grammar
complexity (hence the cost of parsing).
The automatic optimisation of corpus-derived {\sc lscg}s is the
subject of ongoing research and the results reported here for this
method are therefore preliminary.

{\bf Theory Refinement} (ALLiS). ALLiS (\cite{Dejeancoling},
\cite{Dejeanconllb}) is a inductive rule-based system using a
traditional general-to-specific approach \cite{MitchellML}.  After
generating a default classification rule (equivalent to the n-gram
model), ALLiS tries to refine it since the accuracy of these rules is
usually not high enough.  Refinement is done by adding more premises
(contextual elements).  ALLiS uses data encoded in XML, and also
learns rules in XML.  From the perspective of the XML formalism, the
initial rule can be viewed as a tree with only one leaf, and
refinement is done by adding adjacent leaves until the accuracy of the
rule is high enough (a tuning threshold is used).  These additional
leaves correspond to more precise contextual elements.  Using the
hierarchical structure of an XML document, refinement begins with the
highest available hierarchical level and goes down in the hierarchy
(for example, starting at the chunk level and then word level).
Adding new low level elements makes the rules more specific,
increasing their accuracy but decreasing their coverage.  After the
learning is completed, the set of rules is transformed into a
proper formalism used by a given parser.

{\bf Labelled SOM and Memory Based Learning} (LSOMMBL) is a neurally
inspired technique which incorporates a modified self-organising map
(SOM, also known as a `Kohonen Map') in memory-based learning to
select a subset of the training data for comparison with novel
items. The SOM is trained with labelled inputs. During training, each
unit in the map acquires a label. When an input is presented, the node
in the map with the highest activation (the `winner') is identified.
If the winner is unlabelled, then it acquires the label from its
input. Labelled units only respond to similarly labelled
inputs. Otherwise training proceeds as with the normal SOM. When
training ends, all inputs are presented to the SOM, and the winning
units for the inputs are noted.  Any unused units are then
discarded. Thus each remaining unit in the SOM is associated with the
set of training inputs that are closest to it. This is used in MBL as
follows. The labelled SOM is trained with inputs labelled with the
output categories. When a novel item is presented, the winning unit
for each category is found, the training items associated with the
winning units are searched for the closest item to the novel item and
the most frequent classification of that item is used as the
classification for the novel item.

{\bf Maximum Entropy} When building a classifier, one must gather
evidence for predicting the correct class of an item from its
context. The Maximum Entropy (MaxEnt) framework is especially suited
for integrating evidence from various information sources. Frequencies
of evidence/class combinations (called features) are extracted from a
sample corpus and considered to be properties of the classification
process. Attention is constrained to models with these properties. The
MaxEnt principle now demands that among all the probability
distributions that obey these constraints, the most uniform is
chosen. During training, features are assigned weights in such a way
that, given the MaxEnt principle, the training data is matched as well
as possible. During evaluation it is tested which features are {\em
active} (i.e., a feature is active when the context meets the
requirements given by the feature). For every class the weights of the
active features are combined and the best scoring class is chosen
\cite{bdp96}. For the classifier built here we use as evidence the
surrounding words, their POS tags and baseNP tags predicted for the
previous words. A mixture of simple features (consisting of one of the
mentioned information sources) and complex features (combinations
thereof) were used. The left context never exceeded 3 words, the right
context was maximally 2 words. The model was calculated using existing
software \cite{Dehaspe97}.

{\bf Inductive Logic Programming (ILP)} Aleph is an ILP machine
learning system that searches for a hypothesis, given positive (and,
if available, negative) data in the form of ground Prolog terms and
background knowledge (prior knowledge made available to the learning
algorithm) in the form of Prolog predicates. The system, then,
constructs a set of hypothesis clauses that fit the data and
background as well as possible.  In order to approach the problem of
NP chunking in this context of single-predicate learning, it was
reformulated as a tagging task where each word was tagged as being
`inside' or `outside' a baseNP (consecutive NPs were treated
appropriately). Then, the target theory is a Prolog program that
correctly predicts a word's tag given its context. The context
consisted of PoS tagged words and syntactically tagged words to the
left and PoS tagged words to the right, so that the resulting tagger
can be applied in the left-to-right pass over PoS-tagged text.

{\bf Minimum Description Length} (MDL) Estimation using the 
minimum description length principle involves finding a model
which not only `explains' the training material well, but also is
compact. The basic idea is to balance the generality of a model
(roughly speaking, the more compact the model, the more general it is)
with its specialisation to the training material. We have applied MDL
to the task of learning broad-covering definite-clause grammars from
either raw text, or else from parsed corpora
\cite{Osbo99b}. Preliminary results have shown that learning using
just raw text is worse than learning with parsed corpora, and that
learning using both parsed corpora and a compression-based prior is
better than when learning using parsed corpora and a uniform prior.
Furthermore, we have noted that our instantiation of MDL does not
capture dependencies which exist either in the grammar or else in
preferred parses.  Ongoing work has focused on applying random field
technology (maximum entropy) to MDL-based grammar learning (see
Osborne \shortcite{Osbo00a} for some of the issues involved).

{\bf Finite State Transducers} are built by interpreting probabilistic
automata as transducers.  We use a probabilistic grammatical algorithm,
the DDSM algorithm \cite{thollard2001}, for learning automata that
provide the probability of an item given the previous ones.  The items
are described by bigrams of the format feature:class.  In the
resulting automata we consider a transition labeled feature:class as
the transducer transition that takes as input the first part (feature) of
the bigram and outputs the second part (class).  By applying the
Viterbi algorithm on such a model, we can find out the most probable
set of class values given an input set of feature values.  As the DDSM
algorithm has a tuning parameter, it can provide many different
automata.  We apply a majority vote over the propositions made by the
so computed automata/transducers for obtaining the results mentioned
in this paper.

{\bf Memory-based learning} methods store all training data and
classify test data items by giving them the classification of the
training data items which are most similar.  We have used three
different algorithms: the nearest neighbour algorithm {\sc ib1ig},
which is part of the Timbl software package \cite{daelemans99}, the
decision tree learner {\sc IGTree}, also from Timbl, and C5.0, a
commercial version of the decision tree learner C4.5 \cite{Quinlan93}.
They are classifiers which means that they assign phrase classes such
as I (inside a phrase), B (at the beginning of a phrase) and O
(outside a phrase) to words.  In order to improve the classification
process we provide the systems with extra information about the words
such as the previous {\it n} words, the next {\it n} words, their
part-of-speech tags and chunk tags estimated by an earlier
classification process.  We use the default settings of the software
except for the number of examined nearest neighbourhood regions for {\sc
ib1ig} (k, default is 1) which we set to 3.

\subsection{Combination techniques}

When different systems are applied to the same problem, a clever
combination of their results will outperform all of the individual
results \cite{dietterich97}.
The reason for this is that the systems often make different errors
and some of these errors can be eliminated by examining the
classifications of the others.
The most simple combination method is {\sc majority voting}.
It examines the classifications of the test data item and for each
item chooses the most frequently predicted classification.
Despite its simplicity, majority voting has found to be quite useful
for boosting performance on the tasks that we are interested in.

We have applied majority voting and nine other combination methods
to the output of the learning systems that were applied to the three
tasks.
Nine combination methods were originally suggested by Van Halteren et
al. \shortcite{vanhalteren98}.
Five of them, including majority voting, are so-called voting methods.
Apart from majority voting, all assign weights to the predictions of
the different systems based on their performance on non-used training
data, the tuning data. 
{\sc TotPrecision} uses classifier weights based on their accuracy.
{\sc TagPrecision} applies classification weights based on the
accuracy of the classifier for that classification.
{\sc Precision-Recall} uses classification weights that combine the
precision of the classification with the recall of the competitors.
And finally, {\sc TagPair} uses classification pair weights based on
the probability of a classification for some predicted classification
pair \cite{vanhalteren98}.

The remaining four combination methods are so-called {\sc stacked
classifiers}.
The idea is to make a classifier process the output of the 
individual systems.
We used the two memory-based learners {\sc ib1ig} and {\sc IGTree} as
stacked classifiers.
Like Van Halteren et al. \shortcite{vanhalteren98}, we evaluated two
features combinations.
The first consisted of the predictions of the individual systems and
the second of the predictions plus one feature that described the
data item.
We used the feature that, according to the memory-based learning
metrics, was most relevant to the tasks: the part-of-speech tag of the
data item.

In the course of this project we have evaluated another combination
method: {\sc best-N majority voting} \cite{tks2000b}.
This is similar to majority voting except that instead of using the
predictions of all systems, it uses only predictions from some of the
systems for determining the most probable classifications.
We have experienced that for different reasons some systems perform
worse than others and including their results in the majority vote
decreases the combined performance.
Therefore it is a good idea to evaluate majority voting on subsets
of all systems rather than only on the combination of all systems.

Apart from standard majority voting, all combination methods require
extra data for measuring their performance which is required for 
determining their weights, the tuning data.
This data can be extracted from the training data or the training data
can be processed in an n-fold cross-validation process after which the
performance on the complete training data can be measured.
Although some work with individual systems in the project has been
done with the goal of combining the results with other systems, tuning
data is not always available for all results.
Therefore it will not always be possible to apply all ten combination
methods to the results.
In some cases we have to restrict ourselves to evaluating majority
voting only.

\section{Results}

This sections presents the results of the different systems applied to
the three tasks which were central to this this project:
chunking, NP chunking and NP bracketing.

\subsection{Chunking}

Chunking was the shared task of CoNLL-2000, the workshop on
Computational Natural Language Learning, held in Lisbon, Portugal in
2000 \cite{tks2000c}.
Six members of the project have performed this task.
The results of the six systems (precision, recall and F$_{\beta=1}$
can be found in table \ref{tab-chunking}.
Belz \shortcite{belz2001} used Local Structural Context Grammars for
finding chunks.
D\'ejean \shortcite{dejean2000}
applied the theory refinement system ALLiS to the shared task data.
Koeling \shortcite{koeling2000} evaluated a maximum entropy learner
while using different feature combinations (ME).
Osborne \shortcite{osborne2000} used a maximum entropy-based
part-of-speech tagger for assigning chunk tags to words (ME Tag).
Thollard \shortcite{thollard2001} identified chunks with Finite State
Transducers generated by a probabilistic grammar algorithm (FST).
Tjong Kim Sang \shortcite{tks2000d} tested different configurations of
combined memory-based learners (MBL).
The FST and the LSCG results are lower than those of the other systems
because they were obtained without using lexical information.
The best result at the workshop was obtained with Support Vector
Machines \cite{kudoh2000}.

\begin{table}[t]
\begin{center}
\begin{tabular}{|l|c|c|c|}\cline{2-4}
\multicolumn{1}{l|}{}
                           & precision & recall & F$_{\beta=1}$ \\\hline
MBL         & 94.04\% & 91.00\% & 92.50 \\
ALLiS       & 91.87\% & 92.31\% & 92.09 \\
ME          & 92.08\% & 91.86\% & 91.97 \\
ME Tag      & 91.65\% & 92.23\% & 91.94 \\
LSCG        & 87.97\% & 88.17\% & 88.07 \\
FST         & 84.92\% & 86.75\% & 85.82 \\\hline
combination & 93.68\% & 92.98\% & 93.33 \\\hline
best        & 93.45\% & 93.51\% & 93.48 \\
baseline    & 72.58\% & 82.14\% & 77.07 \\\hline
\end{tabular}
\end{center}
\caption{
The chunking results for the six systems associated with the project
(shared task CoNLL-2000).
The baseline results have been obtained by selecting the most frequent
chunk tag associated with each part-of-speech tag.
The best results at CoNLL-2000 were obtained by Support Vector
Machines.
A majority vote of the six LCG systems does not perform much worse than this 
best result.  A majority vote of the five best systems outperforms 
the best result slightly ($5\%$ error reduction).
} 
\label{tab-chunking}
\end{table}

Because there was no tuning data available for the systems, the only
combination technique we could apply to the six project results was
majority voting.
We applied majority voting to the output of the six systems while
using the same approach as Tjong Kim Sang \shortcite{tks2000d}:
combining start and end positions of chunks separately and restoring
the chunks from these results. 
The combined performance (F$_{\beta=1}$=93.33) was close to
the best result published at CoNLL-2000 (93.48).

\subsection{NP chunking}

The NP chunking task is the specialisation of the chunking task in
which only base noun phrases need to be detected.
Standard data sets for machine learning approaches to this task were
put forward by Ramshaw and Marcus \shortcite{ramshaw95}.
Six project members have applied a total of seven different systems to
this task, most of them in the context of the combination paper Tjong
Kim Sang et al. \shortcite{tks2000b}.
Daelemans applied the decision tree learner C5.0 to the task.
D\'ejean used the theory refinement system ALLiS for finding noun
phrases in the data.
Hammerton \shortcite{hammerton2001} predicted NP chunks with the
connectionist methods based on self-organising maps (SOM).
Koeling detected noun phrases with a maximum entropy-based learner
(ME).
Konstantopoulos \shortcite{konstantopoulos2000} used Inductive Logic
Programming (ILP) techniques for finding NP chunks in unseen
texts\footnote{Results are unavailable for the ILP approach.}.
Tjong Kim Sang applied combinations of {\sc ib1ig} systems (MBL) and
combinations of {\sc IGTree} learners to this task.
The results of the six of the seven systems can be found in table
\ref{tab-npchunking}.
The results of C5.0 and SOM are lower than the others because neither
of these systems used lexical information.

\begin{table}[t]
\begin{center}
\begin{tabular}{|l|c|c|c|}\cline{2-4}
\multicolumn{1}{l|}{}
                           & precision & recall & F$_{\beta=1}$ \\\hline
MBL         & 93.63\% & 92.88\% & 93.25 \\
ME          & 93.20\% & 93.00\% & 93.10 \\
ALLiS       & 92.49\% & 92.69\% & 92.59 \\
IGTree      & 92.28\% & 91.65\% & 91.96 \\
C5.0        & 89.59\% & 90.66\% & 90.12 \\
SOM         & 89.29\% & 89.73\% & 89.51 \\\hline
combination & 93.78\% & 93.52\% & 93.65 \\\hline
best        & 94.18\% & 93.55\% & 93.86 \\
baseline    & 78.20\% & 81.87\% & 79.99 \\\hline
\end{tabular}
\end{center}
\caption{
The NP chunking results for six systems associated with the project.
The baseline results have been obtained by selecting the most frequent
chunk tag associated with each part-of-speech tag.
The best results for this task have been obtained with a combination
of seven learners, five of which were operated by project members.
The combination of these five performances is not far off these best results.
} 
\label{tab-npchunking}
\end{table}

For all of the systems except SOM we had tuning data and an extra
development data set available.
We tested all ten combination methods on the development set and
best-3 majority voting came out as the best (F$_{\beta=1}$ = 93.30;
it used the MBL, ME and ALLiS results).
When we applied best-3 majority voting to the standard test set,
we obtained F$_{\beta=1}$ = 93.65 which is close to the best result we
know for this data set
(F$_{\beta=1}$ = 93.86) \cite{tks2000b}.
The latter result was obtained by a combination of seven learning
systems, five of which were operated by members of this project.

The original Ramshaw and Marcus \shortcite{ramshaw95} publication
evaluated their NP chunker on two data sets, the second holding a
larger amount of training data (Penn Treebank sections 02-21) while
using 00 as test data.
Tjong Kim Sang \shortcite{tks2000a} has applied a combination of
memory-based learners to this data set and obtained F$_{\beta=1}$ =
94.90, an improvement on Ramshaw and Marcus's 93.3.

\begin{table}[t]
\begin{center}
\begin{tabular}{|l|c|c|c|}\cline{2-4}
\multicolumn{1}{l|}{}
                           & precision & recall & F$_{\beta=1}$ \\\hline
MBL         & 90.00\% & 78.38\% & 83.79 \\
LSCG        & 80.04\% & 80.25\% & 80.15 \\
MDL         &  53.2\% &  68.7\% &  59.9 \\\hline
best        & 91.28\% & 76.06\% & 82.98 \\
baseline    & 77.57\% & 59.85\% & 67.56 \\\hline
\end{tabular}
\end{center}
\caption{
The results for three systems associated with the project for the
NP bracketing task, the shared task at CoNLL-99.
The baseline results have been obtained by finding NP chunks in the
text with an algorithm which selects the most frequent chunk tag
associated with each part-of-speech tag.
The best results at CoNLL-99 was obtained with a bottom-up
memory-based learner.
An improved version of that system (MBL) delivered the best project
result.
The MDL results have been obtained on a different data set and
therefore combination of the three systems was not feasible.
} 
\label{tab-npbracketing}
\end{table}

\subsection{NP bracketing}

Finding arbitrary noun phrases was the shared task of CoNLL-99, held
in Bergen, Norway in 1999.
Three project members have performed this task.
Belz \shortcite{belz2001} extracted noun phrases with Local Structural
Context Grammars, a variant of Data-Oriented Parsing (LSCG).
Osborne \shortcite{osborne99} used a Definite Clause Grammar learner
based on Minimum Description Length for finding noun phrases in
samples of Penn Treebank material (MDL).
Tjong Kim Sang \shortcite{tks2000a} detected noun phrases with a
bottom-up cascade of combinations of memory-based classifiers (MBL).
The performance of the three systems can be found in table
\ref{tab-npbracketing}.
For this task it was not possible to apply system combination to the
output of the system.
The MDL results have been obtained on a different data set and
this left us with two remaining systems.
A majority vote of the two will not improve on the best system and
since there was no tuning data or development data available, other
combination methods could not be applied.

\section{Prospects}

The project has proven to be successful in its results for applying
machine learning techniques to all three of its selected tasks:
chunking, NP chunking and NP bracketing.  We are looking forward to
applying these techniques to other NLP tasks.  Three of our project
members will take part in the CoNLL-2001 shared
task, `clausing', hopefully with good results.  Two more have started
working on the challenging task of full parsing, in particular by
starting with a chunker and building a bottom-up arbitrary phrase
recogniser on top of that.  The preliminary results are encouraging
though not as good as advanced statistical parsers like those of
Charniak \shortcite{charniak2000} and Collins \shortcite{collins2000}.

It is fair to characterise LCG's goals as primarily technical in the
sense that we sought to maximise performance rates, esp. the
recognition of different levels of NP structure.  Our view in the
project is certainly broader, and most project members would include
learning as one of the language processes one ought to study from a
computational perspective---like parsing or generation.  This suggest
several further avenues, e.g., one might compare the learning progress
of simulations to humans (mastery as a function of experience).  One
might also be interested in the exact role of supervision, in the
behaviour (and availability) of incremental learning algorithms, and
also in comparing the simulation's error functions to those of human
learners (wrt to phrase length or construction frequency or
similarity).  This would add an interesting cognitive perspective to
the work, along the lines begun by Brent \shortcite{brent:97}, but we
note it here only as a prospect for future work.

\subsection*{Acknowledgement}

LCG's work has been supported by a grant from the European Union's
programme {\it Training and Mobility of Researchers}, ERBFMRXCT980237.

\small
\bibliographystyle{acl}

\end{document}